\documentclass[a4paper,10pt]{article}
\usepackage{amsmath}
\usepackage{mathtools}
\usepackage{amsfonts}
\usepackage{amssymb}
\usepackage{multirow} %added
\usepackage{algorithm}
\usepackage{algpseudocode}

\begin{document}

\title{Predator--Prey Model: Driven Hunt for Accelerated Grokking}

\author{I. A. Lopatin,  S. V. Kozyrev, and Alexander N. Pechen
\thanks{
This research was funded by the Ministry of Science and Higher Education of the Russian Federation project 075-15-2024-529.}
\thanks{Steklov Mathematical Institute of Russian Academy of Sciences, Gubkina 8, Moscow 119333, Russia (e-mail: ilya.lopatin2606@gmail.com, kozyrev@mi-ras.ru, apechen@gmail.com).}
}

%\markboth{Journal of IEEE Transactions on Artificial Intelligence, Vol. 00, No. 0, Month 2020}
%{Lopatin \MakeLowercase{\textit{et al.}}: Predator--Prey Model: Driven Hunt for Accelerated Grokking}

\maketitle

\begin{abstract}
A machine learning method is proposed using two agents (representing two coupled replicas of the model parameters) that simulate the biological behavior of a predator and a prey. In this method, the two agents interact with each other via a pair of potentials that simulate the dynamics when the predator chases the prey while the prey runs away from the predator --- to perform an optimization on the landscape. This method allows, for optimization landscapes with narrow ravines with gentle slope or even flat bottoms, to avoid  optimization getting stuck in a ravine.  For some examples of grokking (i.e., delayed generalization) problems we show that this method allows for achieving up to a hundred times faster learning in gradient evaluations and up to 60 times in the wall-clock time compared to the standard learning procedure.
\end{abstract}


\section{Introduction}

Grokking phenomenon (i.e., delayed generalization in overparameterized systems) in learning problems was discovered and investigated in various works including~\cite{Grokking1,Grokking4}. This phenomenon is manifested as follows: first, the model memorizes the data achieving high accuracy (i.e. percentage of correct predictions) only on the training data set while failing to generalize; then, if training is continued (using some version of stochastic gradient descent), a situation of ''delayed generalization'' or ''understanding/awareness'' occurs: the model shows an increase in the accuracy at the test data set. This phenomenon means that learning with generalization is possible for overparameterized systems with relatively small data sets which is considered impossible in the classical machine learning. Grokking has been mostly limited to fairly small or synthetic tasks (such as modular arithmetic and a reduced MNIST setup).
The only case with grokking for real-world data was reported recently~\cite{GrokkRealData}. In \cite{Always}, possible wide presence of grokking was suggested. Examples of grokking were found and delayed robustness as vanishing of adversarial examples was demonstrated for various neural networks with more complex examples such as training of a convolutional neural network (CNN) on CIFAR10 or a Resnet on Imagenette.

In \cite{GrokkingArxiv,ljm2025} the following picture was proposed to explain grokking ---
learning in the presence of grokking is an optimization problem over a ravine landscape, i.e. a landscape containing narrow ravines with flat bottoms. In the optimization process, the system quickly falls into a ravine, and then performs random walk along the ravine. Such random walk is significantly slower compared to gradient descent. Thus, learning gets stuck in the ravine, which explains the delay in the generalization during grokking. Falling into the ravine corresponds to the sample memorization phase during learning, and grokking corresponds to a slow motion along the ravine until reaching the generalization domain. This domain has increased entropy, which entails the irreversibility of the transition to generalization during grokking due to the second law of thermodynamics \cite{GrokkingArxiv,ljm2025}.

The Ravine method for optimization over ravine landscapes was developed by Gelfand and Tsetlin~\cite{Gelfand0,Gelfand}. Closely related Nesterov accelerated gradient optimization method was proposed~\cite{Nesterov}. Both these methods  benefit from fast convergence for general convex objective functions~\cite{Attouch2022}. Modern optimization methods such as Adam and AdamW are modifications of the accelerated gradient methods. All these methods operate efficiently for optimization over landscapes where the slope of the ravine bottom is small but non zero; in this case the motion along the ravine can be accelerated. Different modern variants of the ravine method were developed in~\cite{AdamRAG},~\cite{Attouch2024}.

Ravine landscapes do appear in a variety of optimization phenomena. They are relevant for overparameterized systems, where local minima merge and a zero-risk manifold appears. In this case the bottom of ravines is flat, the gradient along the ravine is zero, and the optimization problem is non-convex.
In \cite{Belkin} it was shown that for  generic optimization landscapes for overparameterized systems, a major part of the parameter space satisfies the Polyak--Lojasiewicz condition which provides the convergence of the gradient descent to vicinity of zero-risk manifold, i.e. to vicinity of ravines. This property describes memorization but not generalization. Only part of the zero-risk manifold corresponds to generalization. Ravine landscapes appear in optimal control of open quantum systems, where for various commonly used quantum control objectives the sub-manifold of the
global maximum solutions was shown to be a connected set forming the bottom of the ravine~\cite{PechenJPA2008}. Quantum control problems are relevant to multiple applications in quantum technologies and machine learning, e.g., with the use of deep learning Alpha Zero~\cite{DalgaardNPJ2020}. The ravine structure of the optimization landscape also appears in a vast range of optimization problems in chemistry and material properties~\cite{MoorePCCP2011}, where machine learning approaches are widely used~\cite{Chibani2020}.
The ravine structure of fitness landscapes in biological evolution with globally
optimal genotypes forming a connected level set of equivalent fitness was predicted in the physically grounded Optimal landscapes in Evolution (OptiEvo) theory~\cite{FengChemSci2011}.

In the present work, the Predator--Prey Model (PPM) (see also~\cite{Eyring,ljm2025}) optimization method for avoiding getting stuck in a ravine is proposed based on the use of two agents, called for convenience the predator and the prey, representing two coupled replicas of the model parameters and interacting with each other via pair of specially adjusted potentials: repulsive potential $P$ for prey and attracting potential $V$ for predator. Both agents evolve according to the dynamics determined by these potentials and by data gradient landscape but with replacing standard gradient descent step by Adam/AdamW algorithm step. We have two versions of the algorithm: one is when both agents have two coupled replicas of the optimization state (Algorithm~\ref{alg:PP-classicPreTrained}), and another when they share the same optimizer (Algorithm~\ref{alg:PP-con}).
For each algorithm we consider two options: when in all epochs predator evaluates data gradient with the same minibatch as prey, or does not evaluate at all.

Biological interpretation is the following: two agents interact not only with the training landscape, but also the prey runs away from the predator while the predator chases the prey. Thus, the predator drives the prey along the ravine. Therefore, we also call this approach the driven hunt method. We show for several examples that the learning acceleration can achieve orders of magnitude.
For landscapes without ravine structure, the method can perform inferior in comparison to standard methods. Also, for the proposed method it is important for the bottom of the ravine to be connected, otherwise it can never reach generalization. For the PPM  for overparameterized systems, transition to generalization is also irreversible, but the mechanism of irreversibility is different in comparison to grokking and has the dynamical nature.

We apply the proposed PPM to well known problems: modular arithmetic (for the transformer architecture) and MNIST handwritten digit recognition using a multilayer perceptron (MLP)~\cite{Grokking1,Grokking4}, and to optimization for the Rosenbrock function. We perform numerical experiments which show that, compared to the original grokking models, the proposed method allows to achieve an acceleration of training up to a hundred times in the number of gradient calls. The average achieved wall clock time acceleration for test launches is found to be about 60 times when a single GeForce RTX 4060 GPU was used.

The proposed method is a bio-inspired  two agent optimization method. There is extensive literature on multiagent and bio-inspired optimization, in particular particle swarm optimization methods~\cite{Swarm}. Typically multiagent methods employ many agents with similar swarm behavior used for exploration of the optimization landscape. Multi--Agent Re\-in\-force\-ment Learning (MARL) was considered~\cite{MARL}.  Also asymmetric (adversarial) methods were investigated, in particular two agent methods such as Generative Adversarial Networks (GANs) \cite{GAN} with the minimax objective, and methods with coupled replicas of parameters such as  BYOL \cite{BYOL}, siamese networks \cite{SimSiam} and VICReg \cite{VICReg} used in Self-Supervised Learning (SSL).  In comparison to GANs, our method is different, the particular form of the agent interaction is important, and the application of the method to ravine landscapes is crucial.
It might happen that the reason for efficiency of operation of algorithms like \cite{BYOL} and \cite{SimSiam} could be clarified by the analogy to the PPM algorithm.

The structure of the paper is the following. In Section \ref{sec:Baseline} we reproduce the known results on grokking for modular arithmetic and MNIST using AdamW algorithm. In Section \ref{sec:PPM} we introduce the PPM  and perform numerical simulations for modular arithmetic and MNIST showing a significant increase in the optimization efficiency. In Section \ref{sec:PP-connModel} we propose the PPM-conn variant of the PPM where the predator and the prey share connected momenta, showing even higher increase in the optimization efficiency. In Section \ref{sec:GrokTimeDataSize} the exponential dependence of the grokking time on the sample size and linear dependence on initial weight norm of model for AdamW algorithm is proven. In Section \ref{sec:Rosenbrock}, we test the PPM for optimization over the landscape of the Rosenbrock function. In Section \ref{sec:Ablation}, the ablation for the PPM parameters is performed. In Section \ref{sec:Regimes} we compare the PPM with other methods. Conclusions Section \ref{sec:Conclusion} summarizes the work. The Appendix Section \ref{app:Models} contains the details of the numerical simulations.

\section{Original grokking problems and standard learning method}\label{sec:Baseline}

Grokking was first demonstrated for the problem of learning for modular arithmetic \cite{Grokking1} modulo $p=97$, in particular, for the division operation $\phi (a, b) = a \cdot b^{-1} ~\mathrm{mod}~ p$. A random subsample of the maximum possible sample (which is the set of all pairs of residues $\mathrm{mod}~p$) was considered as the training sample. The test sample is the remaining part of the maximum possible sample which was not included in the training sample. Decoder--Only Transformer \cite{Vaswani} was considered as the trained model. Below we call this problem as ModuloOperation.
In Sections \ref{sec:Baseline}--\ref{sec:PP-connModel}, \ref{sec:Regimes} ModuloOperation is the division modulo 97 and in Section \ref{sec:GrokTimeDataSize} it is modular addition for $p=139$. In \cite{Grokking4}, one of the considered problems was recognizing handwritten digits on the MNIST dataset using MLP, which below for brevity we call as MNIST.

Detailed experimental schemes for these two problems, the algorithm for creating training and test samples, the exact model architectures and their initialization schemes are described in the Appendix \ref{app:Models}. As a standard training procedure, we consider optimization by the AdamW algorithm \cite{AdamWSource} with a standard set of parameters: training step $\alpha = 10^{-3}$, weight decay $\lambda = 10^{-2}$, exponential moving average coefficients for the first and second moments $b_1 = 0.9$ and $b_2 = 0.999$, respectively, and the additional parameter for the numerical stability $\epsilon = 10^{-8}$.

\begin{figure}[h]
\centering
\includegraphics[width= \linewidth]{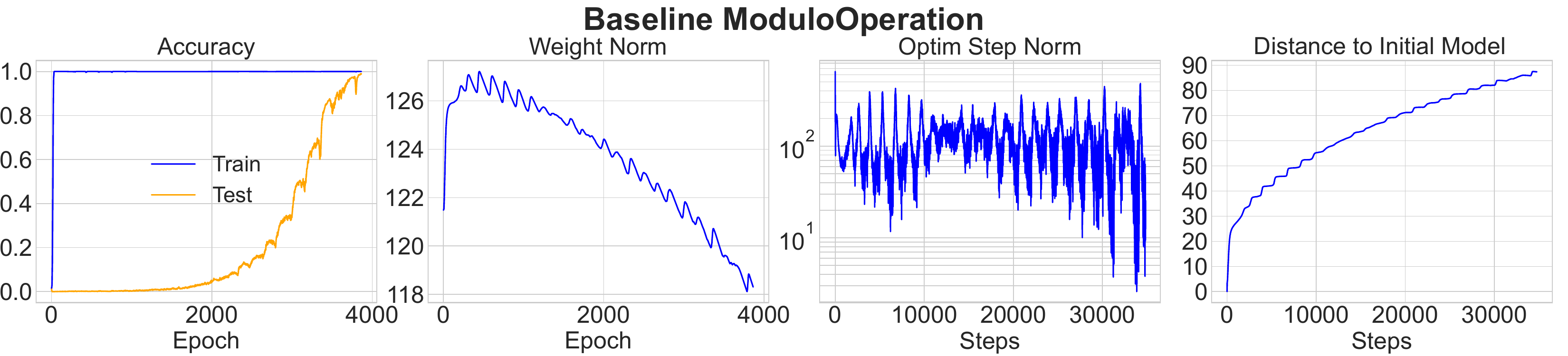}
\includegraphics[width= \linewidth]{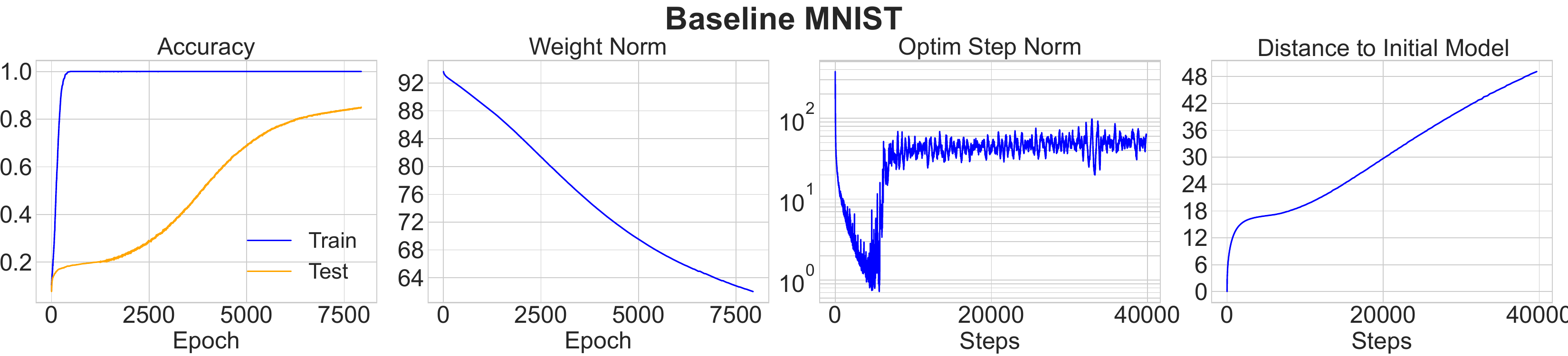}
\caption{Typical training processes for ModuloOperation and MNIST with the standard AdamW algorithm. Upper row: for ModuloOperation. Bottom row: for MNIST. In each row from left to right: accuracy of the training model vs epoch, weight norm of the training model vs epoch, effective optimizer step norm $\|\hat{m}_t / ( \sqrt{\hat{v}_t} + \epsilon)\|$ for Adam/AdamW vs optimization process step, and distance to the initial model vs optimization process step.}
\label{fig:StandardLearning}
\end{figure}

We choose different batch sizes and different accuracy thresholds for ModuloOperation and MNIST problems. We define the ''memorization'' phase as achieving accuracy of more than $99 \%$ on the training data, and the ''understanding'' (or generalization) phase as achieving accuracy of more than $99 \%$ on the test data for ModuloOperation and more than $85 \%$ for MNIST. The batch size for ModuloOperation is 512 and for MNIST is 200. For ModuloOperation the epoch size is 9 steps, and for MNIST the epoch size is 5 steps. Typical learning graphs are shown in Fig. \ref{fig:StandardLearning} for ModuloOperation (upper row) and for MNIST (bottom row).  The plots in the first column demonstrate grokking phenomenon, where we observe rapid increase of the accuracy for the train data and then delayed increase of the accuracy for the test data occur after some number of iterations. The weight norm plot in the second column for ModuloOperation shows the ''memorization'' regime (rapid increase of the weight norm) and then slow grokking behavior during the ''understanding'' regime. For MNIST (second plot, bottom), such behavior is less evident. The third and fourth columns will be used below for comparing the standard approach with the PPM.

We calculate the optimizer step norm and distances to the initial model for each optimization step (i.e. after each batch), but accuracy and weight norm are calculated only for each epoch (i.e. after forward and backward procedure for all bathes).  We do this because the optimizer step has large variations with significant fluctuations from batch to batch, and calculating accuracy among these is the most time-efficient operation.

We represent the set of model parameters by vector in the Euclidean space, and by the model norm and the distance between models we mean the corresponding Euclidean norm and the distance between parameter vectors of the models.  Optim Step denotes the training step; for the  AdamW optimization algorithm \cite{AdamWSource} it is the vector of accumulated momentum $ \hat{m}_t / ( \sqrt{\hat{v}_t} + \epsilon) $, where $\hat{m}_t$ is bias-corrected first moment estimate, $\hat{v}_t$ is bias-corrected second raw moment estimate, and $\epsilon$ is small parameter used for numerical stability.

Repeating the runs, we find that the number of optimizer epochs before reaching the generalization phase is about $4 \times 10^3$ for ModuloOperation and about $8 \times 10^3$ for MNIST. This number is strongly influenced by the initial initialization, in particular by the initial weight norm of the model. Therefore, models with the same architecture but with different initialization can demonstrate significantly different grokking times. Hence, initialization affects time to reach the ravine. In our approach, moving towards bottom of the ravine is realized using standard tools (Adam/AdamW), and the PPM is applied to move in the ravine. All acceleration occurs during optimization in the ravine as shows the distance to starting points in the figures~\ref{fig:PP-classic}, \ref{fig:PP-classicGradPred} \ref{fig:PP-con}, \ref{fig:PPconnAalpha} below.

\section{The Predator--Prey Model}\label{sec:PPM}

The PPM here is implemented as the Algorithm \ref{alg:PP-classicPreTrained} following~\cite{Eyring,ljm2025}.\footnote{The code is available at https://github.com/ilyalopatin26/PpmGrok}
In this algorithm, we introduce two agents (the prey and the predator) which represent two coupled replicas of the model parameters and introduce steps 27--30 in the optimization cycle which correspond to the interaction between the prey and  predator agents.
OPTIM in Algorithm \ref{alg:PP-classicPreTrained} denotes the optimizers of the Adam or AdamW type. OPTIM.step denotes the procedure for updating the optimizer state (i.e., pair of vectors $m_t, v_t$) and for updating the target parameters. The entry OPTIM\_agent.state $\gets$ OPTIM.state (where agent is either the prey or the predator) denotes the procedure for synchronizing the state of the OPTIM\_agent optimizer with the state of the OPTIM optimizer, i.e. equating the number of iterations $t$ and the accumulated moments. Note that, generally speaking, the Algorithm \ref{alg:PP-classicPreTrained}  has more hyper-parameters, for example $b_1, b_2$ for moving averages, but for brevity we omit them.

\begin{algorithm}[t]
\caption{PPM with pre-training for grokking}
\begin{algorithmic}[1]
\State \textbf{Hyperparameters:} $A$ (strength of interact.), $\alpha_p$ (predator rate), $\sigma$ (radius of interact.), $N_d$ (additional prey steps), $\alpha$ (learning rate), GradPred, other optim. params
\State \textbf{Input:} $\theta_0$ \Comment{Start vector}
\Function{P}{$d$} \Comment{Potential func.}
  \State \Return $A \cdot e^{-\frac{d}{\sigma}}$
\EndFunction
\State \textbf{Initialize:} $\theta_t \gets \theta_0$, $t \gets 0$, OPTIM $\gets$ OPTIM($\theta_0$, $t=0$, other optim. params)
\State $y_0 \gets \theta_0$ \Comment{Initial predator vector}
\For{$i = 1$ \textbf{to} $N_d$} \Comment{Do additional iterations for non-zero initial distances}
	\State $t \gets t+1$
	\State $g_t \gets \nabla \mathcal{L}(\theta_{t-1}) $
	\State $\theta_t \gets $ OPTIM.step($\theta_{t-1}$, $g_t$, $t$)  \Comment{Update optim. momenta and make step}
\EndFor
\State $x_0 \gets \theta_t$ \Comment{initial prey vector}
\State OPTIM\_prey $\gets$ OPTIM($x_0$, $t=N_d$,  other params) \Comment{to create separate optim. }
\State OPTIM\_pred $\gets$ OPTIM($y_0$, $t=N_d$,  other params)
\State OPTIM\_prey.state $\gets$ OPTIM.state \Comment{ copy accumulated momenta }
\State OPTIM\_pred.state $\gets$ OPTIM.state
\For{$t=N_d$ \textbf{to} ...}
	\State $g_t \gets \nabla \mathcal{L} (x_{t-1}) $
	\State $x_t' \gets$ OPTIM\_prey.step($x_{t-1}$, $g_t$, $t$)
	\If{GradPred}
		\State $g^p_t \gets \nabla \mathcal{L} (y_{t-1}) $
		\State $y_t' \gets$  OPTIM\_pred.step( $y_{t-1}$, $g^{p}_{t}$, $t$)
	\Else
		\State $y_t' \gets y_t$
	\EndIf
	\State $d_t \gets |x_t' - y_t'|$
	\State $l_t \gets (x_t' - y_t') / d_t $
	\State $x_t \gets x_t' + \big( \alpha \cdot P(d_t) \big) l_t $
	\State $y_t \gets y_t' + \big( \alpha \cdot \alpha_p \big) l_t  $
\EndFor
\end{algorithmic}
\label{alg:PP-classicPreTrained}
\end{algorithm}

For initialization of the model we use the AdamW algorithm. First, we apply random initialization. Second, we apply AdamW to achieve memorization of the training sample. Then the PPM algorithm starts.

To clarify the idea, the Predator--Prey Model (PPM) can be considered as a finite difference scheme for a system of ordinary differential equations, where vector $x$ is the prey and vector $y$ is the predator,
\begin{equation}\label{Prey}
\frac{dx}{dt}=-\nabla \mathcal{L}(x)+\frac{x-y}{|x-y|}P(|x-y|),
\end{equation}
\begin{equation}\label{Predator}
\frac{dy}{dt}=-\nabla \mathcal{L}(y)+\frac{x-y}{|x-y|}V(|x-y|).
\end{equation}
Here $|\cdot|$ denotes Euclidean vector norm, $\mathcal{L}$ is the loss function determining the ravine landscape, $P$ is the potential of the predator--prey interaction, and $V$ is the velocity with which the predator chases the prey. In our consideration the predator--prey interaction is directed along the predator--prey direction determined by the unit vector
$l_t=\frac{x-y}{|x-y|}$, the predator chases the prey with constant velocity $V(d)=\alpha_p$,
and the prey chases away with velocity
\begin{equation} \label{eq:PreyPotential}
    P(d) = A e^{ -\frac{d}{\sigma} },
\end{equation}
where $A>\alpha_p$ is the strength of the predator--prey interaction, $\sigma$ is the effective radius of the predator--prey interaction, and $d$ is the predator--prey distance. Algorithm~\ref{alg:PP-classicPreTrained} can be obtained using a standard finite difference scheme for Eqs.~(\ref{Prey})--(\ref{Predator}) but with replacing standard gradient descent step by Adam/AdamW algorithm step in lines 14-26.

Taking the difference of the equations~(\ref{Prey}) and (\ref{Predator}) leads to
$$
\frac{d(x-y)}{dt}=-\left(\nabla \mathcal{L}(x)-\nabla \mathcal{L}(y)\right)+\frac{x-y}{|x-y|}\left(A e^{-|x-y|/\sigma}-\alpha_p\right).
$$
The first bracket with the difference of the gradients, when  $x$ and $y$ are close points in a ravine, is a small quantity which can be neglected. Hence this equation describes convergence (with some fluctuations) of the predator--prey pair to a constant ravine distance. Therefore, for a ravine landscape, after some transition process, the  predator--prey pair with the ravine distance $d_{*} =\sigma\ln\frac{A}{\alpha_p}$ between them will move in a ravine with constant ravine velocity $\alpha_p$. The ravine distance $d_{*}$ and the ravine velocity $\alpha_p$ can be considered as hyperparameters of the model.

\begin{figure}[t]
\centering
\includegraphics[width = \linewidth]{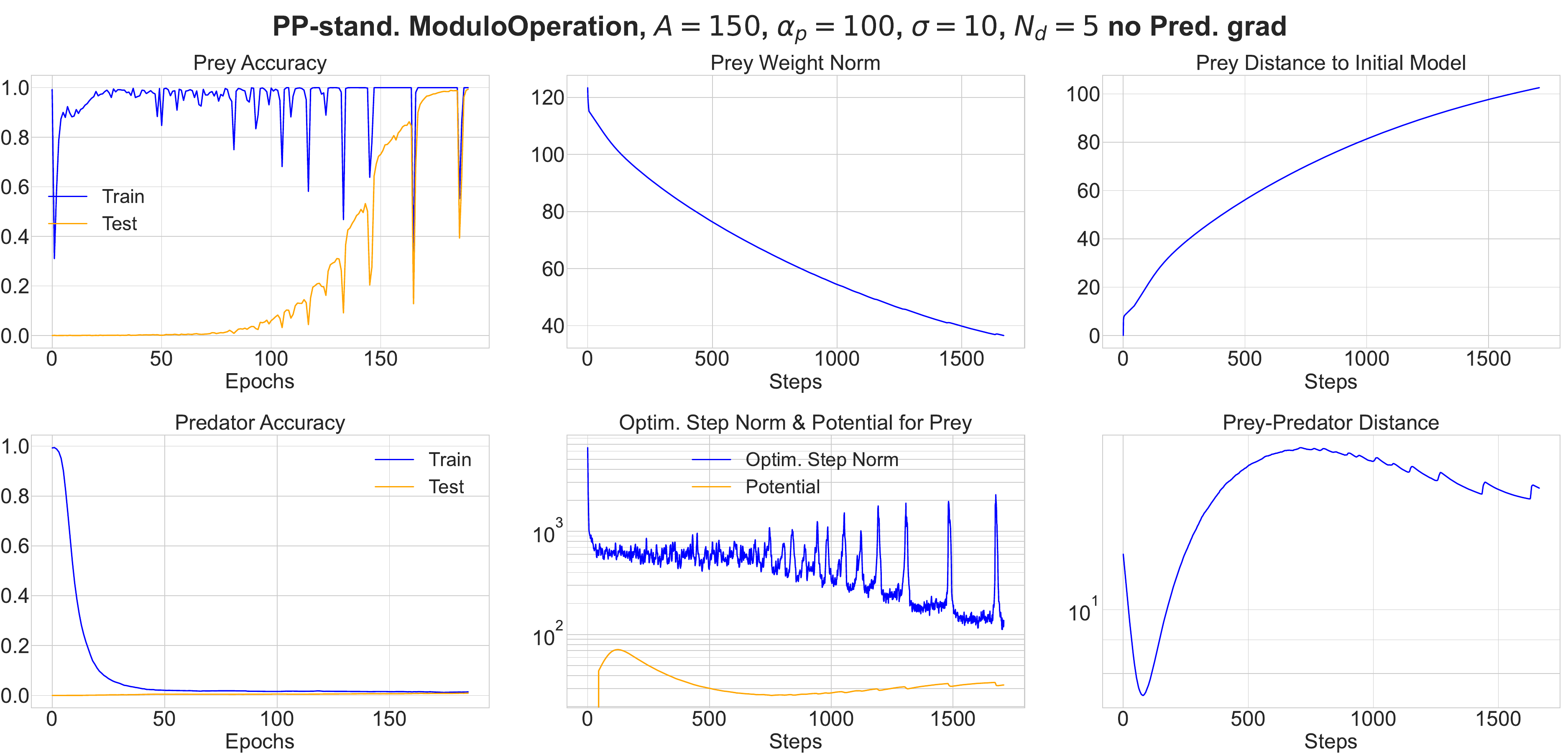}
\includegraphics[width = \linewidth]{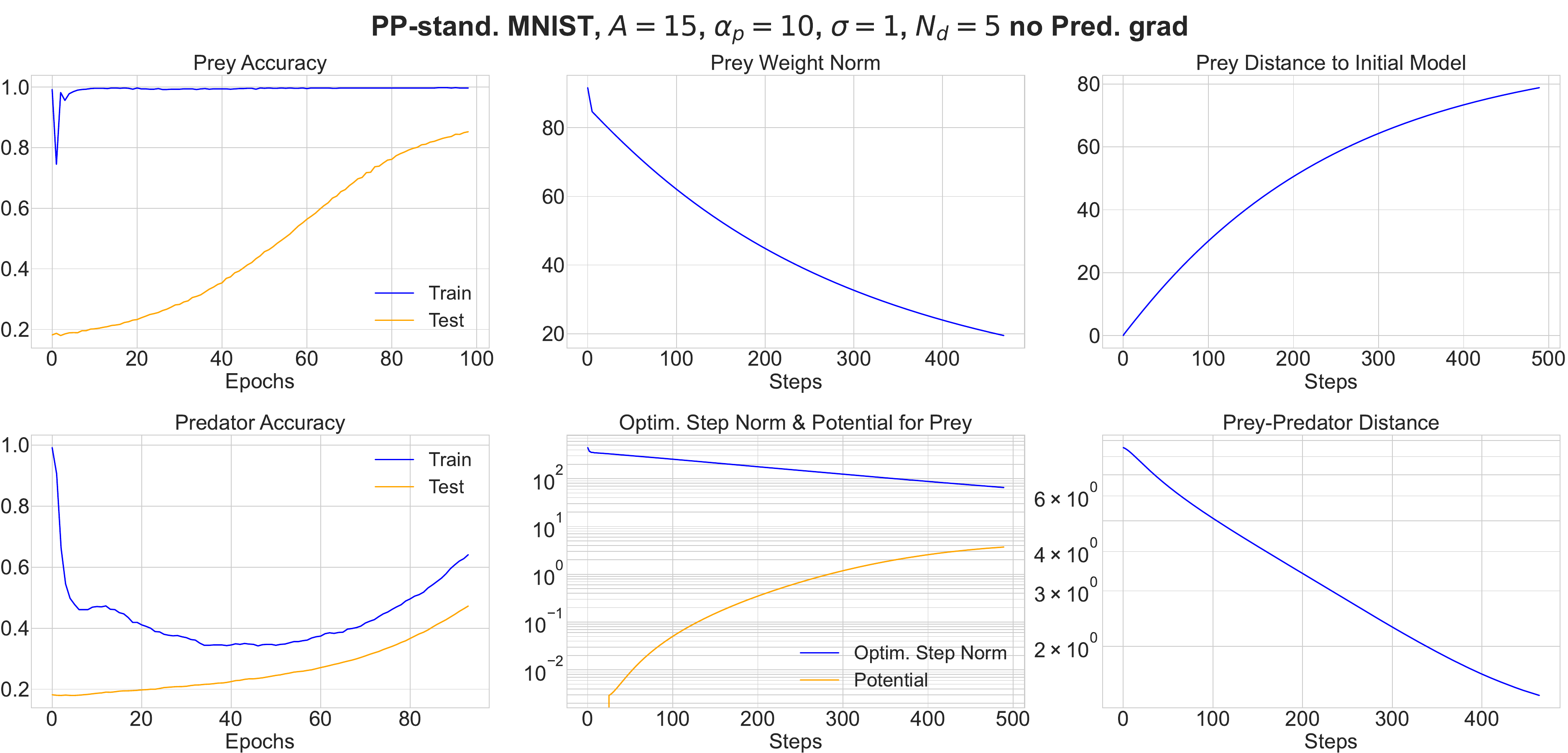}
\caption{Training after the memorization phase using the PPM in the no Pred. grad mode (no gradient calls for the predator) using the Algorithm~\ref{alg:PP-classicPreTrained}. Adam with $\lambda = 10^{-2} $ is used as the optimizer. Two upper rows are for ModuloOperation and two bottom rows are for MNIST. The 1st column is the prey and predator accuracies for train and test data sets over epochs. The 2nd and the 3rd plots in the upper row for each task (i.e. the second and the third rows) represent Prey model weight norm and norm distance between initial Prey model and current Prey model respectively over steps. The 2nd plots in the 2nd and the 4th rows contain comparison of the optimizer step norm $\| \hat{m}_t / (\hat{v}_t + \epsilon)   \|$ for Adam with the prey-predator potential value over steps. We set in (\ref{eq:PreyPotential}) $A=150,\sigma=10$ for ModuloOperation and $A=15, \sigma=1$ for MNIST. For each task we set $\alpha_p=2A/3$ and $N_d=5$.}
\label{fig:PP-classic}
\end{figure}

The results of  training using the Algorithm \ref{alg:PP-classicPreTrained}  are shown in Fig. \ref{fig:PP-classic} for the gradient-off mode, i.e., when GradPred=False in the Algorithm \ref{alg:PP-classicPreTrained} is used so that the gradient for the predator is not called. Hence  one epoch of training in the graphs in Fig. \ref{fig:PP-classic} contains the same number of gradient calls as one epoch in the graphs in Fig. \ref{fig:StandardLearning}. In this case the Algorithm \ref{alg:PP-classicPreTrained}  speeds up the training process by 20 times for ModuloOperation and by 80 times for MNIST in terms of the number of calls to the gradient of the loss function $\mathcal{L}$.

First column in Fig. \ref{fig:PP-classic} shows that the predator loses accuracy and leaves the zero-risk manifold, while the prey does not lose accuracy, perhaps with the exception of the first few iterations and of rare escapes from the ravine followed by fast returns back to the ravine. The effect of the prey leaving the zero-risk manifold at the first iterations can probably be explained as follows. The initial vector between the prey and the predator is almost orthogonal to the zero-risk manifold that forces the prey to move away from this manifold, but then it returns to the ravine due to the resulting effect of the gradient of the loss function $\mathcal{L}$.

\begin{figure}[t]
\centering
\includegraphics[width=\linewidth]{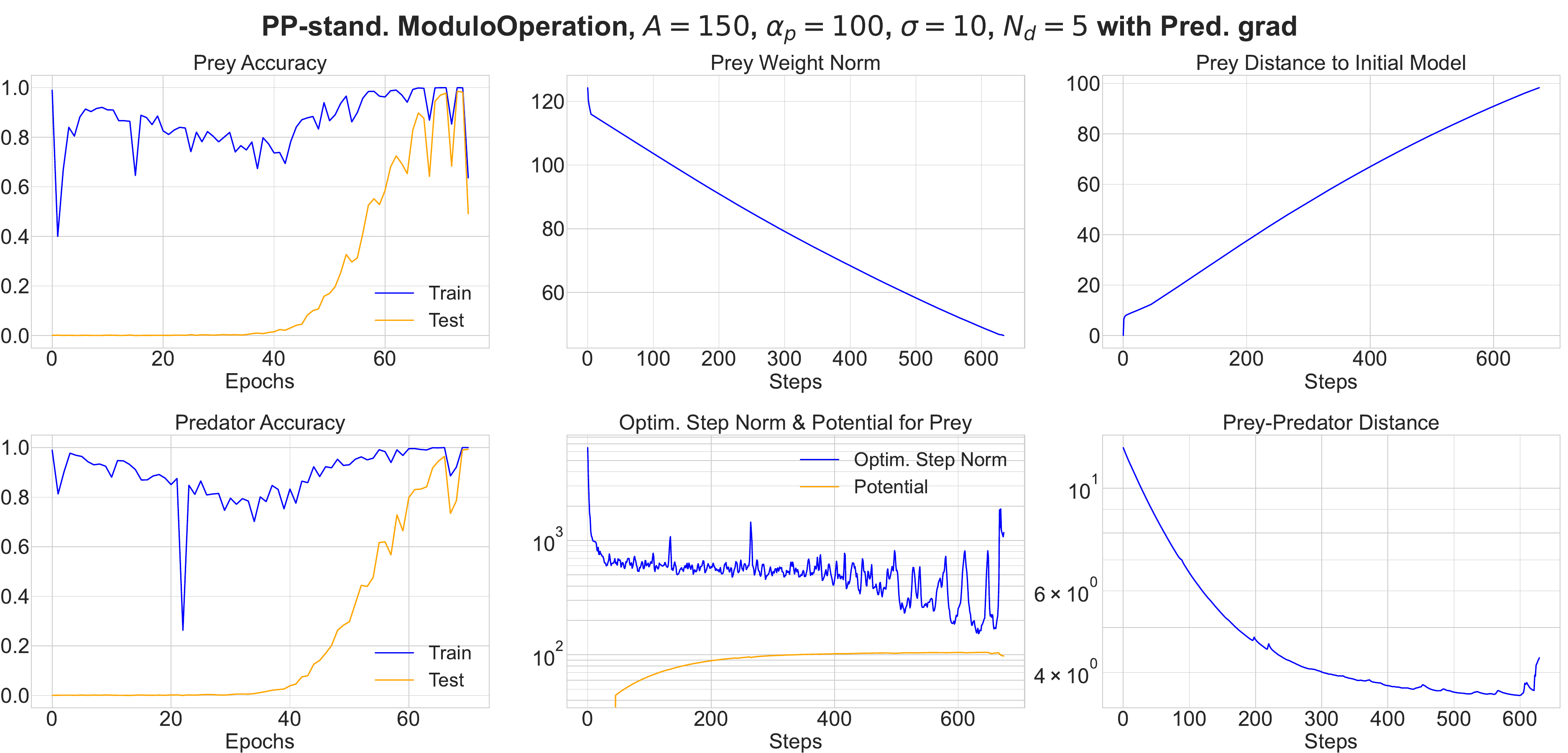}
\caption{A typical example of the PPM standard training process for the ModuloOperation with gradient enabled for the predator. The quantities on the plots are the same as on Fig. \ref{fig:PP-classic}.}
\label{fig:PP-classicGradPred}
\end{figure}

Including in Eq. (\ref{Predator})  the gradient of the loss function at the predator position, $
\nabla \mathcal{L}(y)$, implies that the predator also learns together with the prey, and therefore falls into the ravine of the landscape (reaches the zero-risk manifold) as shown in Fig. \ref{fig:PP-classicGradPred}.

An adversarial pair of agents was considered for GANs~\cite{GAN}, where a minimax objective was used. Our scheme could be formulated as a minimax problem if potentials $P$ and $V$ would be equal, $P=V$, but in our case important is that these potentials depend of the predator--prey distance, are not equal and moreover, have special features.
Therefore, in our method there is no common objective for the two agents even in the minimax form (moreover our approach can not be formulated as minimizing any particular scalar objective).

To determine if a particular problem has grokking or not can be a challenge. We suggest the following procedure. First, start with Adam/AdamW to use it until reach a ravine. If Adam performs well and achieves a desired validation accuracy, we use only it. However, if Adam/AdamW stops to improve test accuracy and overfitting occurs, then we stop Adam/AdamW, and from its stopping point switch to the PPM algorithm. Thus our approach never worse optimization by Adam/AdamW if Adam/AdamW works well, but can improve optimization when Adam/AdamW stops working.

\section{The Predator--Prey Model with Connected Momenta} \label{sec:PP-connModel}

In this section, we present a modified version of the PPM algorithm demonstrating even higher efficiency. In this modified version, we use an Adam-like optimization method, but instead of the two optimizers as in Algorithm \ref{alg:PP-classicPreTrained} we use a single optimizer, and both the predator and the prey have access to its state --- the gradient for the prey is used to compute the gradient for the predator. This approach reduces the number of gradient calls twice compared to the case where gradients are computed for both the predator and the prey.

\begin{algorithm}[t]
\caption{PP-conn.}
\begin{algorithmic}[1]
\State \textbf{Hyperparameters:}  $\alpha$, $\lambda$, $b_1, b_2$, $\epsilon$, $A, \alpha_p, \sigma$, $N_d$, use\_m, GradPred
\State \textbf{Input:} $\theta_0$

\Function{P}{$d$} \Comment{Potential func.}
  \State \Return $A \cdot e^{-\frac{d}{\sigma}}$
\EndFunction
\State $m_t \gets 0$, $v_t \gets 0$, $t \gets 0$ \Comment{Initial momenta}
\Procedure{UpdateAdam}{$p$, $g$, $t$}
	\State $g \gets  g + \lambda  p $ \Comment{Adam-style weight decay}
	\If{use\_m}
		\State $m_t \gets b_1 m_{t-1} + (1-b_1) g $
	\Else
		\State $m_t \gets g$
	\EndIf
	\State $v_t = b_2 v_{t-1} + (1-b_2) g^2$
	\State $\hat{m}_t \gets m_t / (1-b_1^t)$
	\State $\hat{v}_t \gets v_t / (1-b_2^t)$
	\State $p \gets p - \alpha \hat{m}_t / ( \sqrt{\hat{v}_t} + \epsilon ) $
	\State \Return $p$
\EndProcedure

\State \textbf{Initialize:} $\theta_t \gets \theta_0$
\State $y_t \gets \theta_0$ \Comment{initial predator vector }
\For{$t=1$ \textbf{to} $N_d$}
	\State $g \gets \nabla \mathcal{L} (\theta_{t-1})$
	\State $\theta_t \gets$ UpdateAdam($\theta_{t-1}$, $g$, $t$)
\EndFor
\State $x_t \gets \theta_t$ \Comment{initial prey vector }
\For{$t=N_d+1$ \textbf{to} ...}
	\State $g \gets \nabla \mathcal{L} (x_{t-1})$
	\State $x_t' \gets$ UpdateAdam($x_{t-1}$, $g$, $t$)
	\If{GradPred}
		\State $g^p \gets \nabla \mathcal{L} (y_{t-1}) $
	\Else
		\State $g^p \gets 0$
	\EndIf
	\State $y_t' \gets$ UpdateAdam($y_{t-1}$, $g^p$, $t$)
	\State $d_t \gets |x_t' - y_t'|$
	\State $l_t \gets (x_t' - y_t') / d_t $
	\State $x_t \gets x_t' + \big( \alpha \cdot P(d_t) \big) l_t $
	\State $y_t \gets y_t' + \big( \alpha \cdot \alpha_p \big)l_t  $
\EndFor
\end{algorithmic}
\label{alg:PP-con}
\end{algorithm}

\begin{figure*}[ht!]
\includegraphics[width = \linewidth]{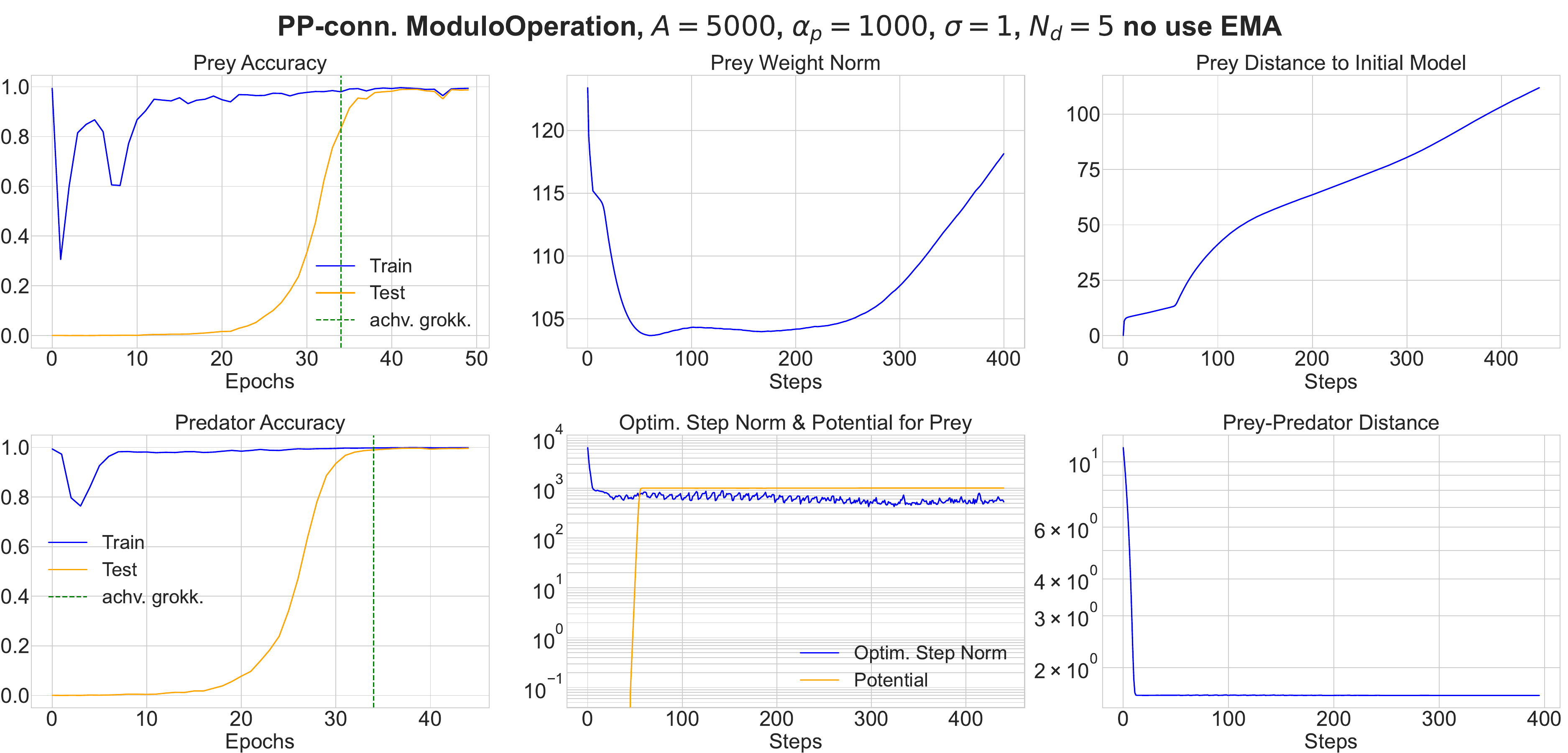}
\includegraphics[width = \linewidth]{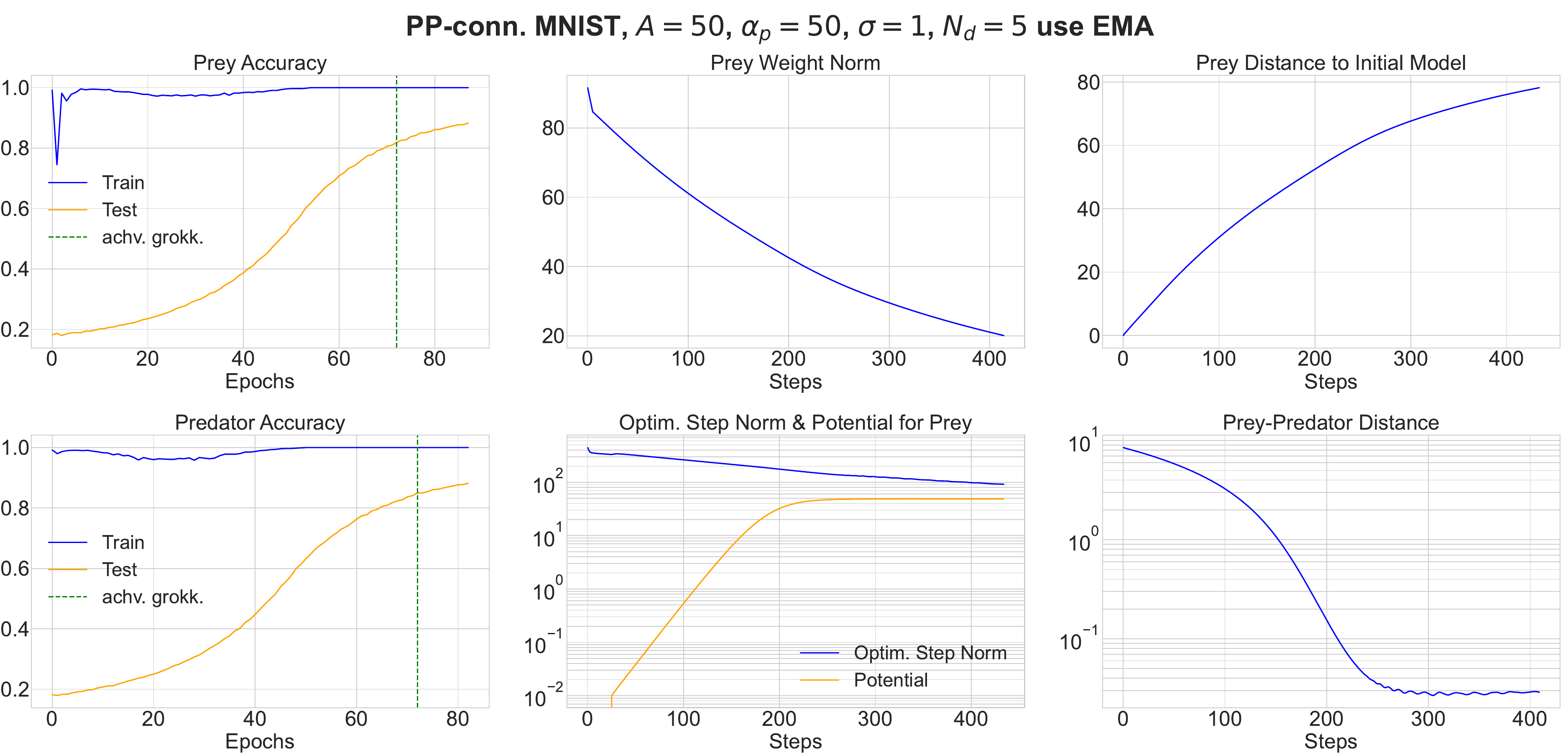}
\caption{Training with PP-conn. In both cases we do not call the gradient for the predator. For the MNIST we use exponential moving average (EMA) for the first moment, for the ModuloOperation we do not use it. The quantities on the plots are the same as on Fig. \ref{fig:PP-classic}. Comparing with the standard procedure shown in Fig. \ref{fig:StandardLearning}, PP-conn achieves an acceleration of training up to a hundred times in the number of gradient calls. We perform additional optimizer epochs after achieving grokking phase to show that accuracy does not decrease, as shown in the left column after the vertical dashed green line.}
\label{fig:PP-con}
\end{figure*}

The results of training with the  Algorithm \ref{alg:PP-con} are shown in Fig. \ref{fig:PP-con}. Since the gradient for the predator is not called, for both ModuloOperation and for MNIST we get an acceleration of grokking of about 100 times in the number of iterations.

\begin{figure}[h]
\includegraphics[width = 21pc]{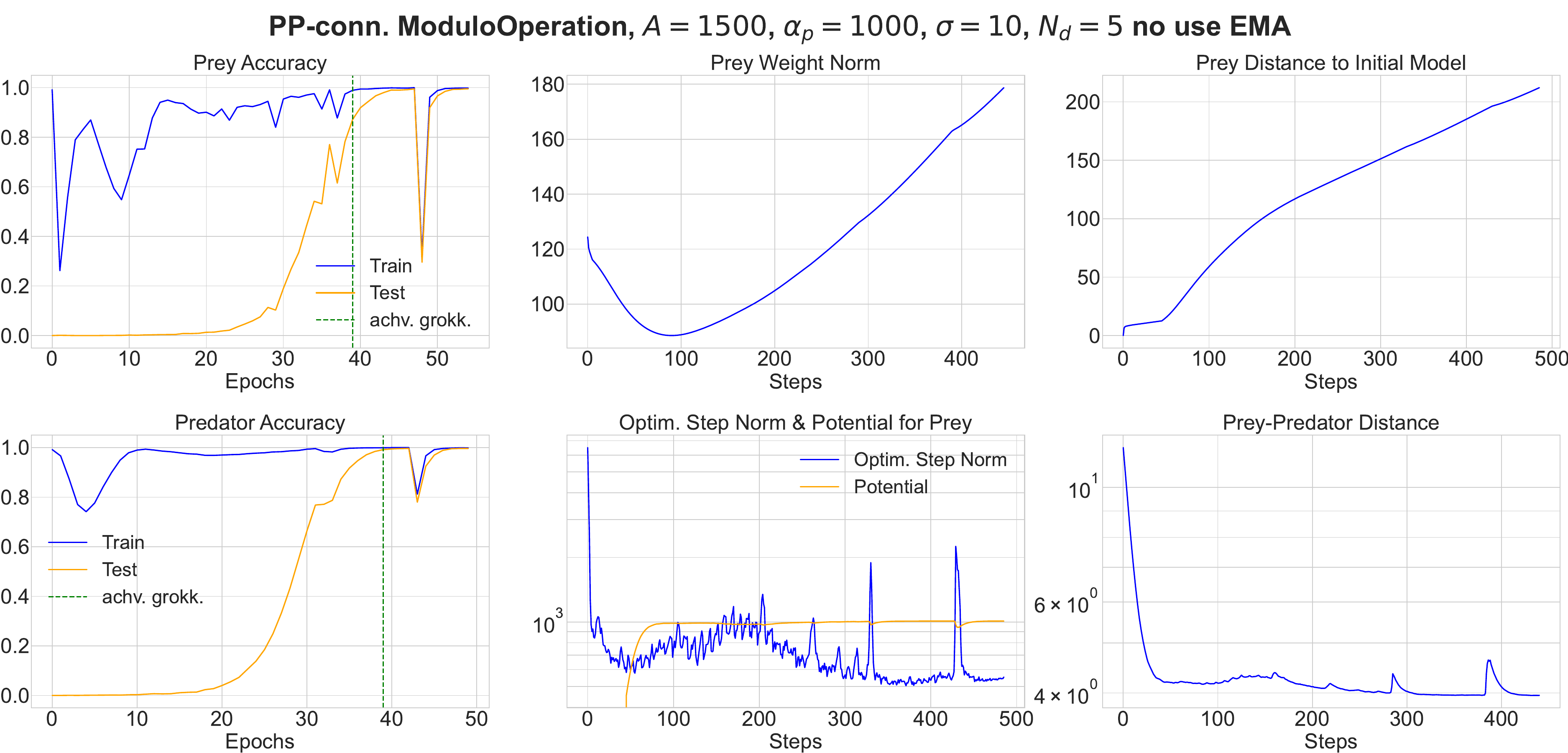}
\includegraphics[width = 21pc]{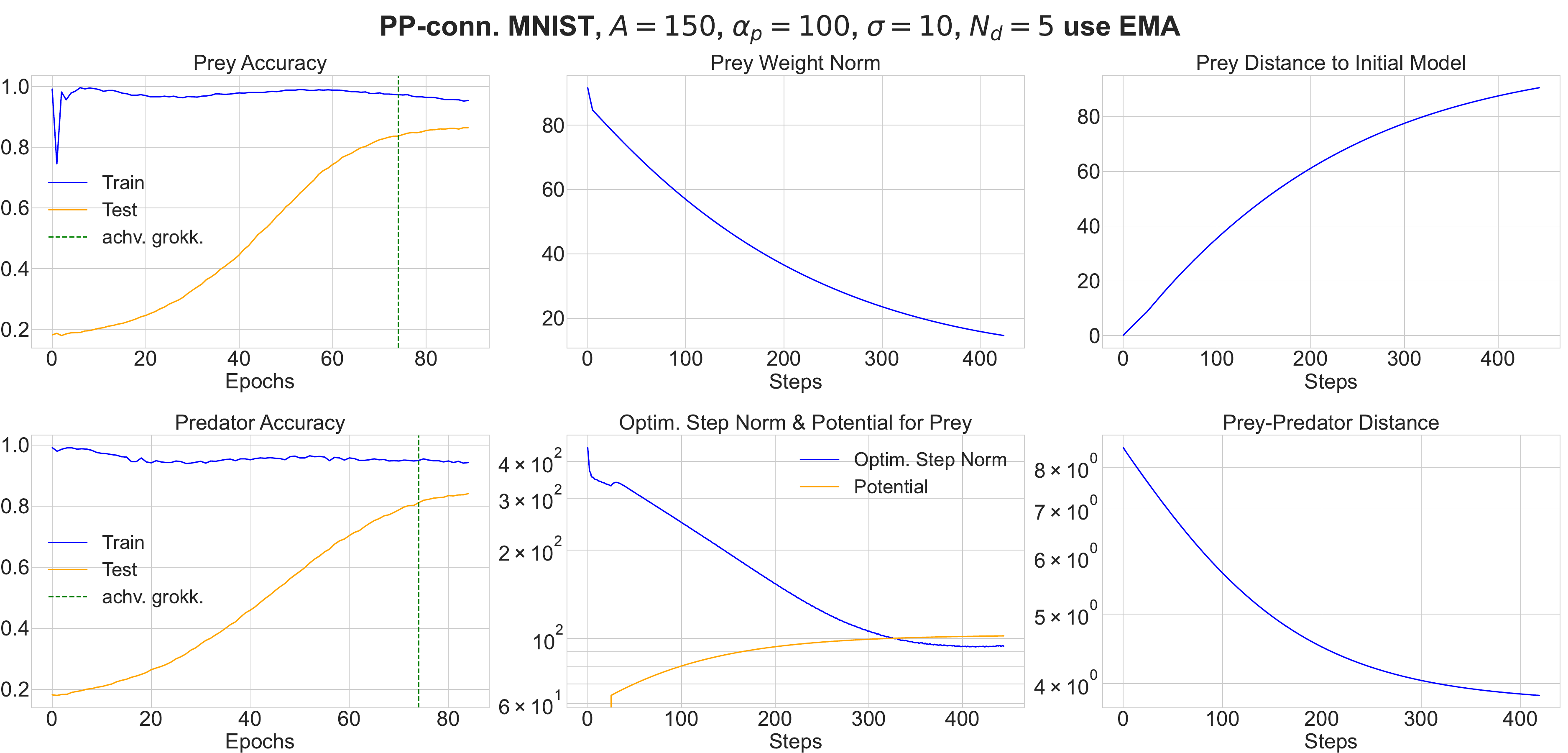}
\caption{Training for PP-conn with $A > \alpha_p$.   The quantities on the plots are the same as on Fig. \ref{fig:PP-classic}.}
\label{fig:PPconnAalpha}
\end{figure}

We note that for this algorithm the potential of the predator--prey interaction tends to a limit value. Running the training for $A > \alpha_p$, one can see in Fig. \ref{fig:PPconnAalpha} that the limiting value of the potential is $\alpha_p$. This behavior is associated with the stabilization of the predator--prey distance which tends to the constant limit value $ d_{*} =\sigma \ln (A/\alpha_p)$, where $P(d_*) =\alpha_p$. Comparing figures \ref{fig:PP-con} and \ref{fig:PPconnAalpha} shows that a wide range of acceptable potential parameters allows to speed up training.
In figures \ref{fig:PP-con} and \ref{fig:PPconnAalpha}  (1st column) we perform additional epochs after achieving the threshold test accuracy to show that there is no accuracy decreasing with the exception of rare escapes from the ravine followed by fast returns to it.

Comparing Fig. \ref{fig:StandardLearning} with Figs. \ref{fig:PP-classic}, \ref{fig:PP-classicGradPred}, \ref{fig:PP-con} and \ref{fig:PPconnAalpha} shows that the predator-prey interaction makes the prey to move in the ravine much more rapidly. For the PP-Conn algorithm the predator influence can significantly exceed the step of optimizer, but it leads to accelerated grokking, as demonstrated on Figs. \ref{fig:PP-con}, \ref{fig:PPconnAalpha} for ModuloOperation.

\section{Dependence of the grokking time on the sample size and on the initial weight norm} \label{sec:GrokTimeDataSize}

In the works \cite{GrokkingArxiv,ljm2025} it was shown that the grokking time decreases exponentially with the growth of the training sample. This behavior can be explained by the description of grokking as a random walk and the application of the Eyring formula of kinetic theory.

In this section, we study numerically the dependence of the grokking time on the data size and on the initial weight model norm. The experimental setup is as follows. Take the prime number $p=139$. To speed up the computations we reduce the size of the neural network (the exact parameters are given in the Appendix) and consider the operation of modular addition. Then the maximal size of the training and test samples, in the case of equal splitting of all data between them, is $S = (p^2-1) / 2 = 9660 $. For $\beta \in (0, 1]$ we take a random subsample of size $\beta S$ and perform a numerical experiment as described in Section \ref{sec:Baseline}.
The initial neural network is the same for all cases.

For the analysis of the grokking time dependence on the initial weight model norm we use a general approach developed for the investigation of optimization effort vs initial data on various landscapes and which was applied to quantum control in~\cite{PechenIsrJChem2012,VolkovPRA2025}. In this approach, the dependence of the  efficiency of optimizing over a landscape (either a ravine or a rugged with multiple local optima) on the initial model is investigated by averaging the optimization effort over uniformly generated random samples in hypercubes of various initial model vector norms.

\begin{figure}[h]
%\centering
\includegraphics[ width =  \linewidth]{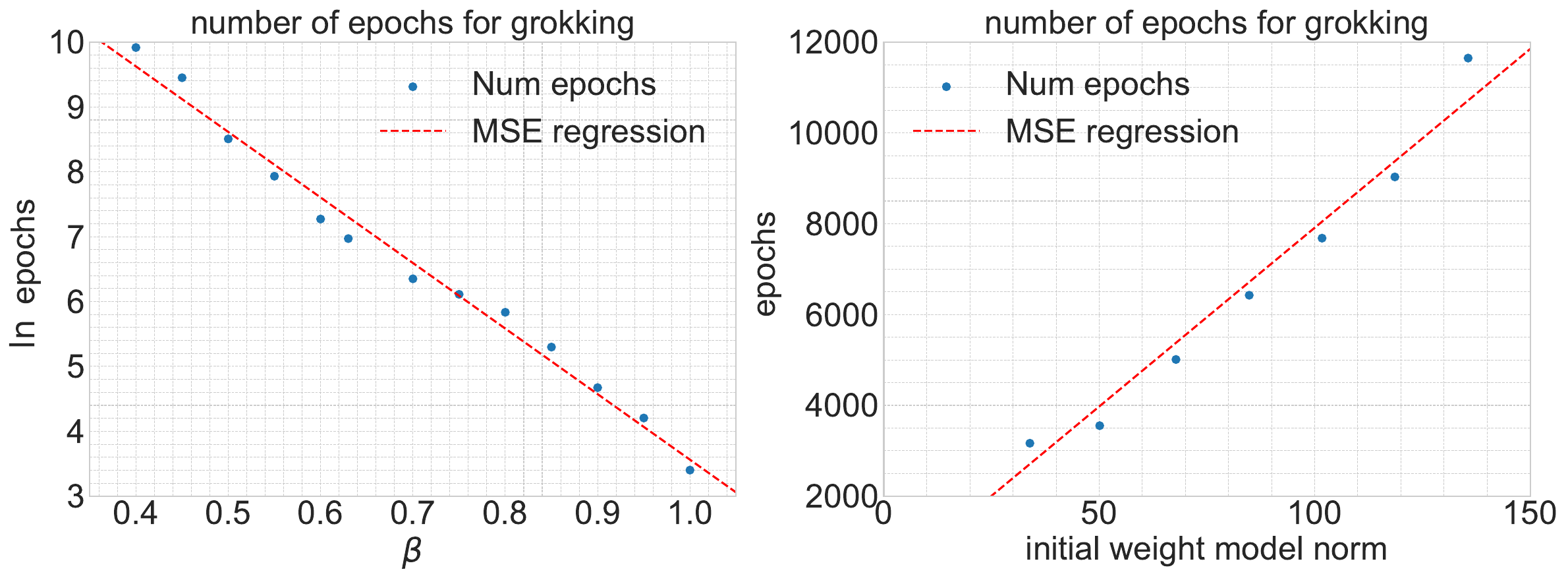}
\caption{Grokking time dependence for addition modulo 139. Left: (in logarithm of the number of epochs) on the sample size $\beta$, which is the proportion of training and test data compared to the maximal possible sample. It shows linear decrease with $\beta$. Right: on the initial weight model norm. It shows linear growth with the weight model norm. Each point is obtained by averaging over 3 runs. }
\label{fig:GrokTimeAlpha}
\end{figure}

The obtained dependencies are shown in Fig. \ref{fig:GrokTimeAlpha}. The results show that the obtained values are well approximated by straight lines. Thus we obtain the exponential law for the time necessary to achieve grokking (in iterations) $T_g \sim B e^{-\frac{\beta}{\eta}}$, where $B$ and $\eta$ are some coefficients. For the analysis of the grokking time dependence on the initial weight model norm we use a moderate averaging over 3 runs for each point which already demonstrates a good linear behavior shown in the right subplot of Fig.~\ref{fig:GrokTimeAlpha}.

For the PPM-conn with Algorithm \ref{alg:PP-con} we have not been able to find an explicit dependence of the number of grokking iterations on the sample size.

\section{Optimization for Rosenbrock function} \label{sec:Rosenbrock}

In this section we compare Adam and PPM-conn on the Rosenbrock function, which is a common testing example with a ravine landscape. For the experiment, we consider an $N$-dimensional version of the Rosenbrock function with the following explicit form, $x=(x_1,\dots,x_N)\in \mathbb{R}^N$
\begin{equation}
    f(x) = \sum_{i=1}^{N-1}\left[ (1-x_i)^2 + 100 (x_{i+1} - x_i^2)^2\right].
\end{equation}
Minimum of this function is $(1,\dots,1)$. The standard initial optimization point $x_0 = (-1.2, 1.0, 1.0, \dots, 1.0)$ is used. We choose $N=50$ with default  learning rate $10^{-3}$ without weight decay. To ensure convergence near the minimum, we use the standard schedule method. We use the following schedule: for each  $S$ steps we update the predator--prey potentials $P$ and $V$ by changing  $A \to A / 2$ and $\alpha_p \to \alpha_p / 2$. In experiments we set $S=200$. In addition to this schedule, we introduce reboots after each $S$ steps by clearing (i.e., setting to zero) the parameters $m,v,t$ for the PPM-conn. For Adam, such reboots were found not to provide any reasonable increase in the performance.

\begin{figure}[h]
    \centering
    \includegraphics[width= 0.9 \linewidth]{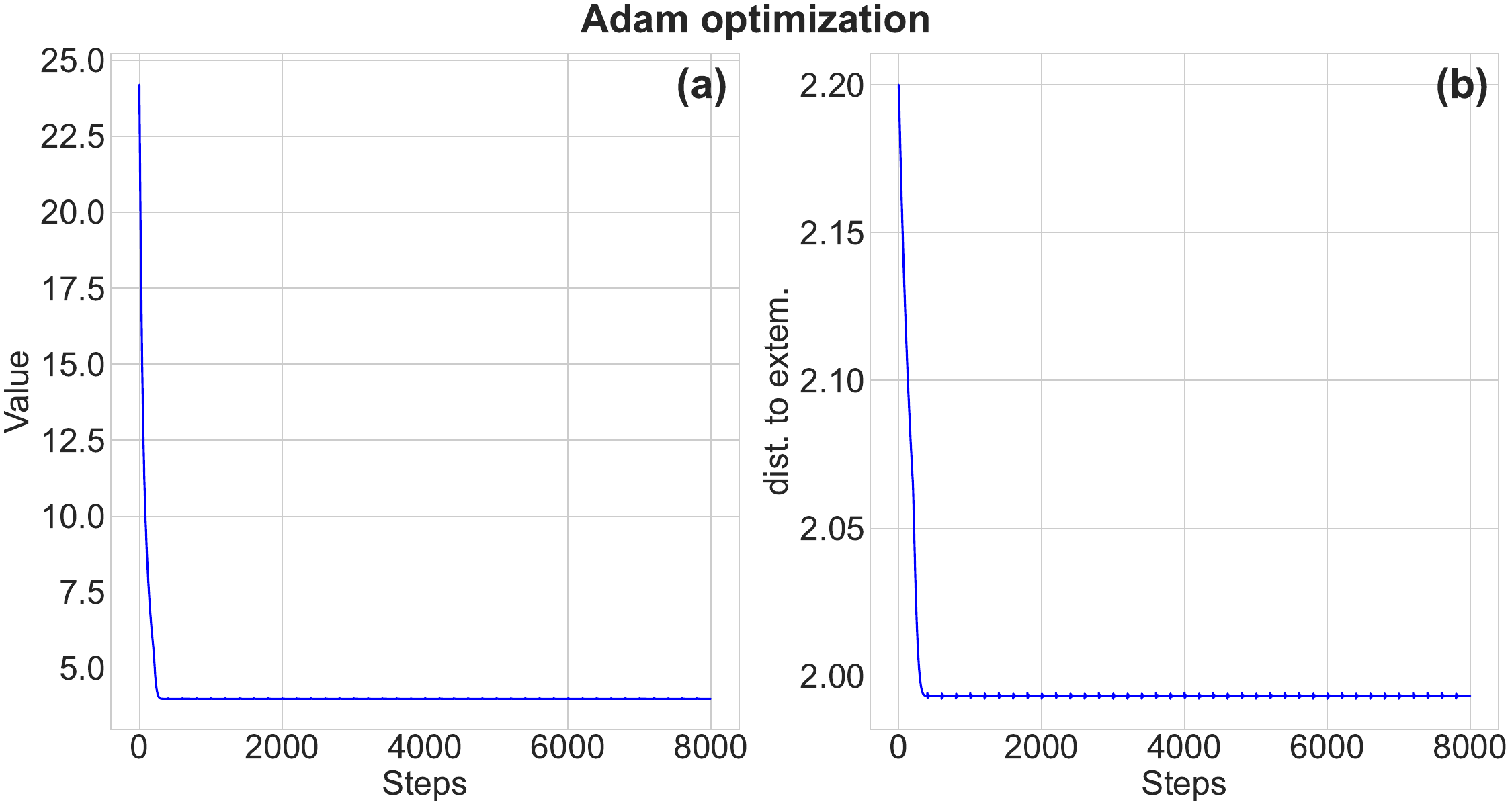} \\
    \includegraphics[width= 0.9 \linewidth]{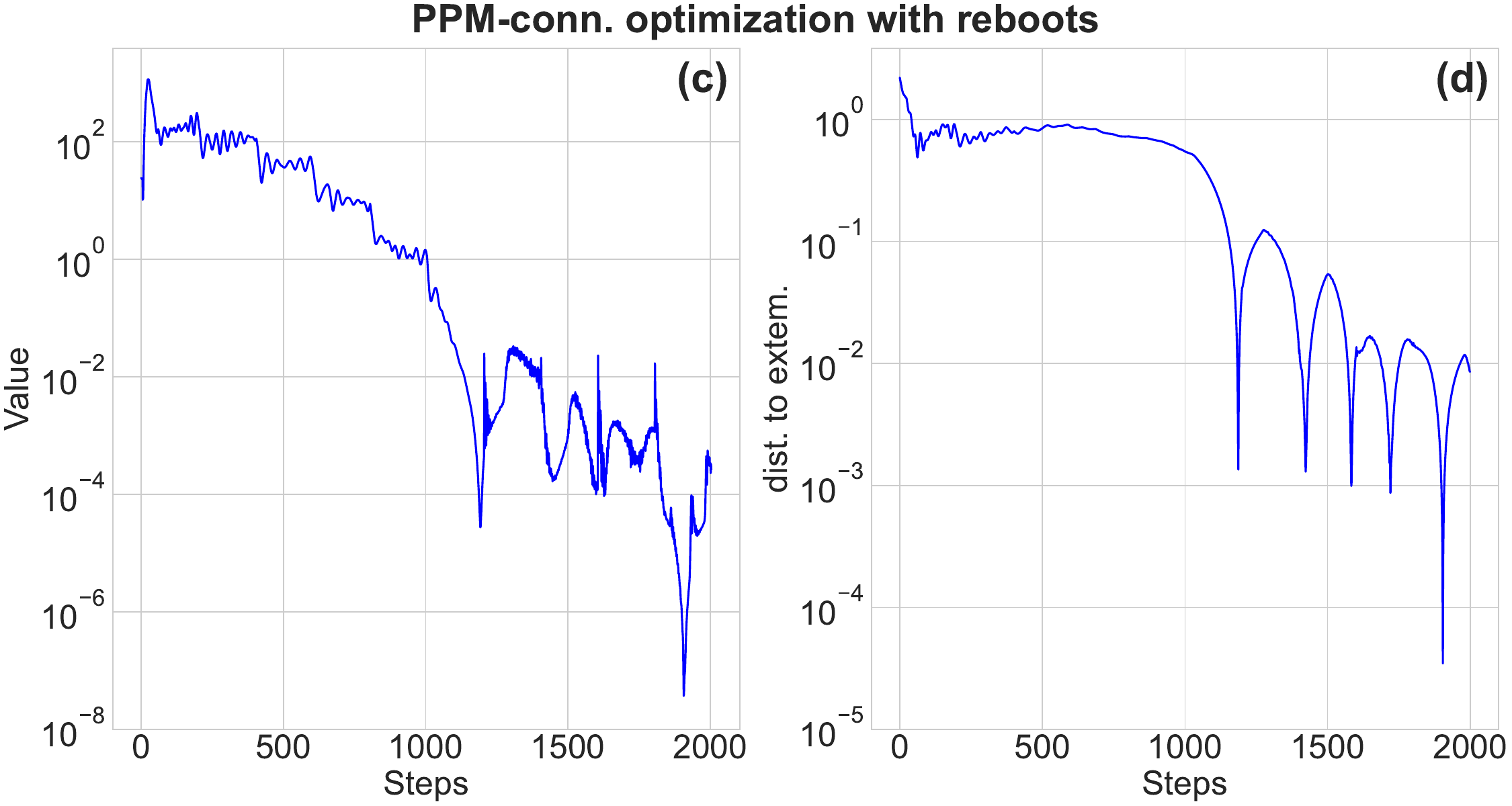}
    \caption{ Comparing optimization process for standard Adam algorithm  and PPM-conn with optimizer reboots. Upper and bottom rows show  Rosenbrock function value (left column) and distance to the minimum (right column). Subplots (a) and (b) for Adam algorithm (with reboots, but for Adam without reboots holds the same trajectory) and subplots (c) and (d) for PPM-conn with reboots.}
    \label{fig:Rosenbrock}
\end{figure}

The results of the numerical simulations with Adam and PPM-conn are shown on Fig.~\ref{fig:Rosenbrock}, upper and bottom rows, respectively. The figures show loss function value (left column) and distance to the minimum (right column). On subplots (a) and (b), there is a fast decay of the loss function value (from about 25 to about 5) and of the distance to the extremum (minimum) in several hundreds of steps and then very slow decrease of both these quantities. Therefore Adam quickly reaches the ravine and then gets stuck there.

On subplots (c) and (d), in about of 1000 steps the PPM-conn reaches the bottom of the ravine, but then it sufficiently fast moves towards the minimum. Both the loss function value and the distance to the minimum on average rapidly decrease. At some steps, they  experience jumps with increase of their values which are due to the normalization of the gradient after the reboots in line 17 of the Algorithm \ref{alg:PP-con} (which is the same as line 12 in the Algorithm 2 of~\cite{AdamWSource}). Important is that after these jumps the algorithm quickly returns to lower values.

However, experimentally, it was found that it is even more effective not to update the parameters, but to completely reboot the PPM optimizer with new reduced parameters.  Note that, the reboot method has no significant influence for Adam procedure in this case. The optimization processes for baseline Adam and PPM-conn with Algorithm \ref{alg:PP-con} shown in Fig. \ref{fig:Rosenbrock}. indicate that  Adam procedure rapidly achieves the ravine where the optimization gets stuck, so that the target value decreases very slowly. Contrary, PPM-conn algorithm does not get stuck and continues its rapid motion along the ravine. The reboot method was proven to be particularly effective.

\section{Ablation}\label{sec:Ablation}

In this section we perform ablation for the PPM algorithm. For brevity, we consider only PPM-conn with Algorithm~\ref{alg:PP-con}. Note that the number of epochs for achieving grokking varies significantly from launch to launch and with different seeds. Even if we start with the same initialized model, we have different training samples that seriously affects the learning trajectory, especially for over-parameterized systems. Therefore, we can only get an approximate estimate of the number of epochs for achieving grokking.  We consider three different values ($1.0, 10.0, 100.0$)  for the effective radius $\sigma$ of the predator-prey interaction $P(d)$. For experiments we use a single GeForce RTX 4060 GPU.  Ablation results for ModuloOperation are provided in Table \ref{tab:AblationMO} and for MNIST in Table \ref{tab:AblationMNIST}. The number $N_d$ of the additional prey steps for the initial distance in Algorithm \ref{alg:PP-con} is fixed as $N_d=5$. Since the predator moves close to the prey fast enough, choosing different small enough values of $N_d$ does not significantly affect the process. To get reasonable statistical averaging results, the data in each cell of Tables~\ref{tab:AblationMO} and~\ref{tab:AblationMNIST} are obtained after averaging over 250 launches of the algorithm.
The method is not very sensitive to hyperparameters, see Tables~\ref{tab:AblationMO} and~\ref{tab:AblationMNIST} --- even when the hyperparameters vary by the order of magnitude the method still works for all these values of  hyperparameters.

\begin{figure}
    \includegraphics[width= 0.9 \linewidth]{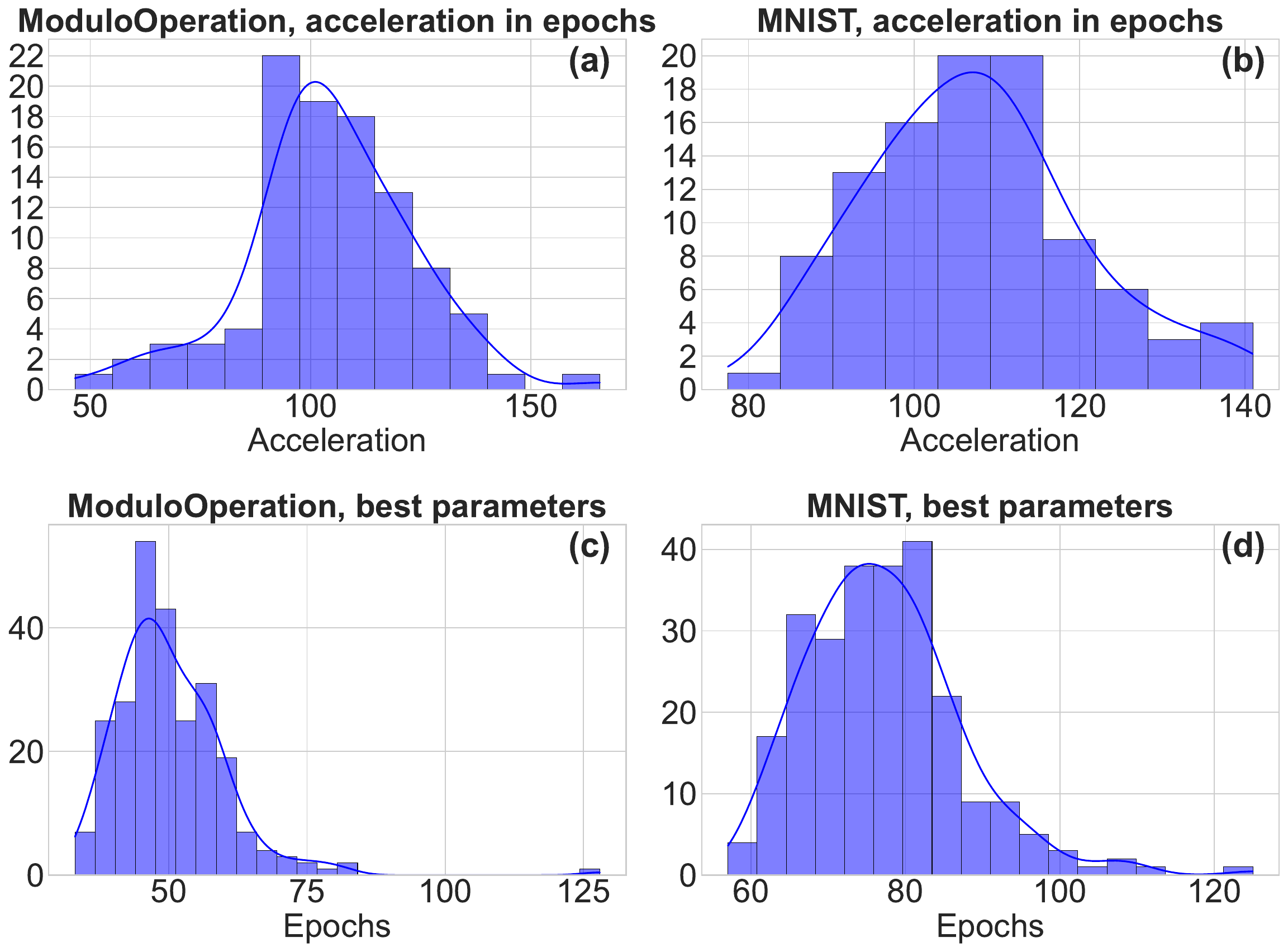}
    \caption{Histograms for ablation. Upper row shows acceleration in epochs for PPM-conn compared with standard AdamW for (a) ModuloOperation and (b) MNIST. Bottom row shows the number of epochs for achieving grokking with PPM-conn for (c) ModuloOperation and (d) MNIST. Parameters for PPM-conn have been chosen as best parameters according to ablation procedure, i.e. tables \ref{tab:AblationMO} and \ref{tab:AblationMNIST} for ModuloOperation and MNIST tasks respectively.}
    \label{fig:Histograms}
\end{figure}

As a result, we find that the best in the considered range PPM-conn parameters for ModuloOperation  are $\sigma=1, A = 5000, \alpha = 1000$ and for MNIST are $\sigma=1, A = \alpha = 50$.  For these parameters,  Fig.~\ref{fig:Histograms} shows the histograms of the number of epochs for achieving grokking and of acceleration for comparing with the standard AdamW procedure. The number of launches for PPM-conn model is 250 and the number of launches for estimation of the average acceleration is 100. The average acceleration in the numbers of epochs and the execution wall-clock time are {\bf 101.6} and {\bf 88.0} for ModuloOperarion and {\bf 106.4} and {\bf 58.5} for MNIST, respectively.

In comparison with AdamW, memory overhead for the PPM-conn in our implementation is 1.8 times more. However, it is not a challenge for the algorithm since the memory overhead for the PPM-conn can be further optimized when necessary by not saving gradient for the predator when it is not called.

\begin{table*}
\caption{Ablation for Modulo Operation for various values of the parameters of PPM-conn.}
\centering
\begin{tabular}{|c|c|c|c|c|c|c|}
\hline
$\sigma$ & $A\backslash\alpha_p$ & 100 & 500 & 1000 & 2000 & 5000 \\
\hline
\hline
\multirow{7}{*}{1} & 100 & 1.00, 107.12 (15.48) & x & x & x & x \\
\cline{2-7}
& 200 &  1.00, 104.72 (13.37) & x & x & x & x \\
\cline{2-7}
& 500 &  1.00, 106.75 (16.88) &  1.00, 90.94 (9.87) & x & x & x \\
\cline{2-7}
& 1000 &  1.00, 110.05 (16.19) & 1.00, 71.43 (7.92) &  1.00, 102.15 (11.30) & x & x \\
\cline{2-7}
& 2000 &  1.00, 113.79 (15.85) &  1.00, 67.00 (8.64) &  1.00, 62.09 (10.03) & 0.98, 185.37 (43.19) & x \\
\cline{2-7}
& 5000 &  1.00, 119.31 (17.54) &  1.00, 75.13 (10.61) & {\bf 1.00, 50.17 (10.14)} & -- &  0.99, 150.29 (36.52) \\
\cline{2-7}
& 10000 &  1.00, 123.81 (17.60) &  1.00, 84.92 (14.82) &  1.00, 57.07 (11.91) & -- &  0.79, 231.94 (46.82) \\
\hline
\hline
\multirow{4}{*}{10} & 100 &  1.00, 116.63 (14.42) & x & x & x & x \\
\cline{2-7}
& 200 &  1.00, 153.95 (22.66)  & x & x & x & x \\
\cline{2-7}
& 500 &  0.97, 205.51 (28.91) &  1.00, 69.07 (8.37) & x & x & x \\
\cline{2-7}
& 1000 &  0.95, 208.21 (30.40) & -- &  1.00, 63.89 (11.71) & x & x \\
\hline
\hline
\multirow{3}{*}{100} & 100 & 0.99, 152.98 (20.02) & x & x & x & x \\
\cline{2-7}
& 200 &  0.91, 226.80 (30.88) & x & x & x & x \\
\cline{2-7}
& 500 & -- &  1.00, 112.15 (22.52) & x & x & x \\
\cline{2-7}
\hline
\multicolumn{7}{l}{}\\[-5pt]
\multicolumn{7}{@{}p{40pc}@{}}{\hspace*{9pt}  The first column shows the values of the parameter $\sigma$. The second column shows the values of the parameter $A$. Upper row shows values of the parameter $\alpha_p$. In cells, we show average result for PPM-conn model with the respective parameters in form ''fraction of success launches, mean number of epochs for achievement grokking, (standard deviation of the number of epochs) ''.  The symbols ''-'' means that procedure do not converge for a significant part (more than $0.75$) of launches in budget of 300 epochs. The marker ''x'' means that $\alpha_p > A$. The data in each cell are obtained after averaging over 250 launches of the algorithm.}\\
\end{tabular}
\label{tab:AblationMO}
\end{table*}

\begin{table*}
\caption{Ablation for MNIST for various values of the parameters of PPM-conn. }
\centering
\begin{tabular}{|c|c|c|c|c|c|c|}
\hline
$\sigma$ & $A\backslash\alpha_p$ & 10 & 50 & 100 & 500 & 1000 \\
\hline
\hline
\multirow{6}{*}{1} & 10 &  1.00, 80.30 (10.35) & x & x & x & x \\
\cline{2-7}
& 50 & 1.00, 88.48 (13.20) & {\bf  1.00, 77.17 (10.03)} & x & x & x \\
\cline{2-7}
& 100 &  1.00, 92.28 (14.82) &  1.00, 101.63 (15.86) &  1.00, 112.99 (20.51) & x & x \\
\cline{2-7}
& 200 &  1.00, 99.76 (19.51) &  1.00, 128.66 (27.36) &  0.98, 175.34 (37.48) & x & x \\
\cline{2-7}
& 500 &  0.95, 131.08 (36.49) &  0.90, 191.33 (41.03) & -- &  1.00, 114.88 (17.96) & x \\
\cline{2-7}
& 1000 & 0.77, 185.72 (46.96) & -- & -- & -- & 1.00, 119.08 (18.92) \\
\hline
\hline
\multirow{3}{*}{10} & 10 &  1.00, 80.92 (10.41) & x & x & x & x \\
\cline{2-7}
& 50 &  0.98, 116.66 (31.65) &  1.00, 79.44 (9.68) & x & x & x \\
\cline{2-7}
& 100 &  0.91, 184.35 (50.54) &  0.77, 214.53 (45.52) &  1.00, 85.07 (10.72) & x & x \\
\hline
\hline
\multirow{3}{*}{100} & 10 &  1.00, 81.02 (10.46) & x & x & x & x \\
\cline{2-7}
& 50 & -- &  1.00, 83.55 (10.90) & x & x & x \\
\cline{2-7}
& 100 & -- & -- &  1.00, 104.13 (16.46) & x & x \\
\hline
\multicolumn{7}{l}{}\\[-5pt]
\multicolumn{7}{@{}p{40pc}@{}}{\hspace*{9pt} \hspace{10pc} Ablation for MNIST task. The notations are the same as in Table \ref{tab:AblationMO}.} \\
\end{tabular}
\label{tab:AblationMNIST}
\end{table*}

As a result, we observe that best outcomes for all the cases are for the same $\sigma=1$.
It agrees with the expectation that for small values of $\sigma$ the method reduces to SGD, while for large values of $\sigma$ (about 100) the prey escapes the ravine and method crushes. Second observation is that small variations around $\sigma=1$ do not significantly affect the optimization; only changes on the orders of magnitude can significantly affect it. Third observation is that most stable for convergence is the choice $A=\alpha_p$ (but optimal parameters can differ, as for ModuloOperation). Summarzing, to tune the method on new problems one could start with $\sigma\sim 1$ and $A=\alpha_p$.

\section{Comparison}\label{sec:Regimes}

In this section, the results of the launch of several standard optimization algorithms from torch.optim library: NAdam, RAdam for two setups: with default parameters and with lr=$10^{-3}$, weight decay=$10^{-2}$; RMSProp, SGD with default parameters, and Lion with the parameters ${\rm lr}=10^{-3}$, ${\rm wd}=10^{-2}$ for both considered tasks are provided. As a result, all these algorithms achieve grokking in about $10^3-10^4$ epochs or even in a longer time as shown in Table~\ref{tab:ComparingOptims}.

\begin{figure}
    \centering
    \includegraphics[width= \linewidth]{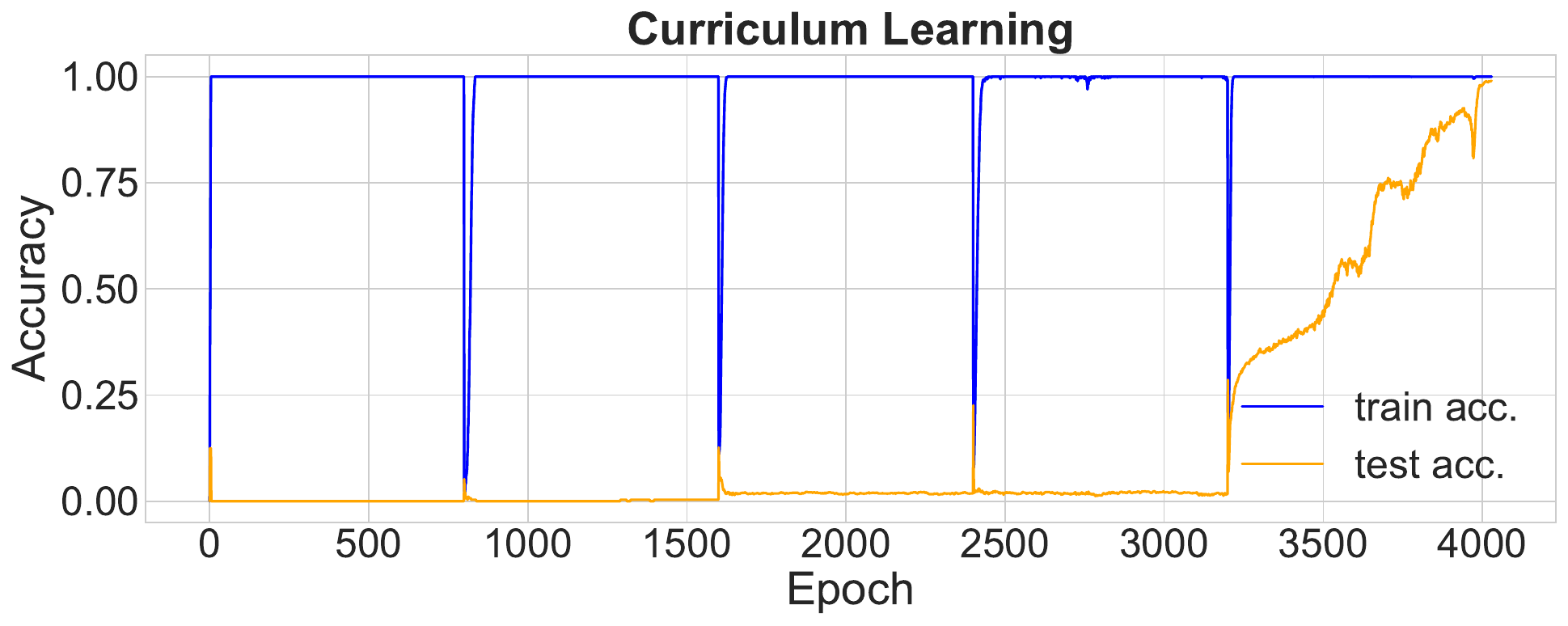}
    \caption{Curriculum learning for modular division task. Sudden drops in train accuracy are caused by changing $p_j$.  }
    \label{fig:Curriculum}
\end{figure}

We also perform with our benchmark  curriculum learning, which was applied for grokking in modulo operation task~\cite{Curriculum}. Initially, we build a model with maximum vocabulary size, for $p=97$ and modular division operation there are 96 possible residues and two symbols for $=$, $*$. Training data contain strings $a*b=c$ with $a, b < p_j$. During training $p_j$ increases to $p$ with the scheduler $p_j\in [5, 25, 49, 73, 97]$, where $p_j$ changes linearly in time each $800$ epochs,  where the maximal number $4000$ of epochs  coincides with the approximate grokking time for baseline as in sec. \ref{sec:Baseline} and Fig. \ref{fig:StandardLearning}. In this approach, we obtain acceleration of grokking wall clock time for the modular division task approximately by factor of two, as shown in Fig. \ref{fig:Curriculum}, in comparison to the standard procedure in sec. \ref{sec:Baseline}. In both cases (Figs.~\ref{fig:StandardLearning} and~\ref{fig:Curriculum}) we use AdamW optimizer with the default parameters. Besides linear curriculum scheduler, we tested other  schedulers from \cite{Curriculum} and a non-fixed prime module or increasing vocabulary size model, but for them we were not able to detect acceleration of grokking.

\begin{table}[h]
    \caption{  Comparing different optimizers}
    \centering
    \begin{tabular}{|c|c|c|}
        \hline
        Optimizer & Modulo Operation & MNIST \\
        \hline
        NAdam, default params. & $2 \cdot 10^3 $ & $1 \cdot 10^4$ \\
        NAdam, lr$=10^{-3}$, wd$=10^{-2}$ & ---  & $8 \cdot 10^2$ \\
        RAdam, default params. & $5 \cdot 10^3$  & --- \\
        RAdam, lr$=10^{-3}$, wd$=10^{-2}$ & --- & $9 \cdot 10^2$ \\
        Lion, lr$=10^{-3}$, wd$=10^{-2}$ &  $4 \cdot 10^3$ & --- \\
        RMSprop, default params. & --- & --- \\
        SGD, default params. & --- & --- \\
        \hline
    \multicolumn{3}{l}{}\\[-5pt]
    \multicolumn{3}{@{}p{21pc}@{}}{\hspace*{9pt}
    The appoximate number of epochs to achieve grokking for ModuloOperation and MNIST. In cells with "---" the procedure does not achieve accuracy test threshold for optimization budget $10^4$ epochs. These computations are time consuming due to long learning. Since explicit numbers for us not necessary and only order of the magnitude is relevant, each number in this Table is obtained after averaging only over five runs which is sufficient for our purposes.}\\
    \end{tabular}
    \label{tab:ComparingOptims}
\end{table}

\section{Conclusions}\label{sec:Conclusion}

In this paper, we introduce a new learning algorithm based on the predator--prey model considered in \cite{Eyring,ljm2025}. This algorithm can be successfully applied to ravine landscapes and is based on the idea of accelerating progress along a ravine using an interaction between two agents, the predator and the prey. The algorithm uses the information about the ravine (predator--prey) direction, that is possible due to application of a multi-agent system.  For two well known model problems, namely learning for modular arithmetic and MNIST, as well as for Rosenbrock function representing a standard ravine landscape, the learning (grokking) speed is found to be accelerated up to a hundred of times in the number of gradient calls.

Even more efficient form of the PPM algorithm, called PPM--conn, is also developed, where the prey and predator adaptive momenta are synchronized. In this algorithm the prey motion is driven by gradient of the loss function, and the predator gets the same driving as the prey that allows for a better synchronization of the predator and the prey motions. Addition of the second agent to the optimization procedure increases the time cost of one epoch and the amount of used memory by the factor of two, but at the same time leads to a significant decrease by two orders of the magnitude for the number of epochs necessary for grokking. For example, in the considered tests averaged acceleration in the wall-clock time is estimated by 60 times, and in the gradient calls by 100 times.

\section*{Appendix}\label{app:Models}

In this Appendix, the detailed setup of the experiments is provided.

For ModuloOperation, we trained the  transformer \cite{Vaswani} with the parameters $d_m=128$, $n = 4$, $l=2$, $d_f = 512$.  In section \ref{sec:GrokTimeDataSize} we used the parameters: $d_m =128$, $n = 2$, $l=1$, $d_f = 256$, $p=139$ with modular addition operation. For linear layers we used common for transformers Xavier initialization, for embedding --- normal initialization with std=1.0, in section \ref{sec:GrokTimeDataSize} std = 0.02 to speed up training. In section \ref{sec:GrokTimeDataSize} we reduce the accuracy threshold to $95 \%$. This is because when data is reduced, the accuracy starts to fluctuate greatly when reaching large values, since the error in prediction on one element changes the percentage ratio more strongly. To reduce the impact of this stochasticity, we lower the acceptance threshold, leaving it large enough to consider that the model has entered the ''understanding'' phase.

For MNIST, a multilayer perceptron with one hidden state was trained (the number of inputs and outputs of the hidden state was set to 200), and the ReLU function was taken as the nonlinearity between the layers. The default uniform initialization was used. In \cite{Grokking4}, the following steps were taken to achieve the grokking effect in this problem, which we repeat in our work. 1) Instead of the full training sample of 60,000 images, a subsample of 1,000 images was taken. 2) Instead of CrossEntropy, which is usually used in classification, the squared deviation of the MSE was taken as the loss function. 3) softmax was not used at the output layer for mapping to the probability space. 4) After initialization, the model weights were additionally multiplied by a specified coefficient in order to increase the weight norm of the original model.

\end{document}